\documentclass[letterpaper]{article} 
\usepackage{aaai2026}  
\usepackage{times}  
\usepackage{helvet}  
\usepackage{courier}  
\usepackage[hyphens]{url}  
\usepackage{graphicx} 
\usepackage{natbib}  
\usepackage{caption} 
\usepackage{amsmath}
\usepackage{algorithm}
\usepackage{algorithmic}
\usepackage{xcolor}
\usepackage{booktabs}
\usepackage{pgfplots}
\usepackage{pgfplotstable}
\usepackage{subcaption}
\usepgfplotslibrary{groupplots}  
\usepackage{multirow}
\usepackage{makecell}
\usepackage[table]{xcolor}
\usepackage{tabularx}
\usepackage{array}

\usepackage{newfloat}
\usepackage{listings}
\DeclareCaptionStyle{ruled}{labelfont=normalfont,labelsep=colon,strut=off} 
\floatstyle{ruled}
\newfloat{listing}{tb}{lst}{}
\floatname{listing}{Listing}

\title{The Finer the Better: Towards Granular-aware Open-set Domain Generalization}

\author{
    Yunyun Wang\textsuperscript{\rm 1},
    Zheng Duan\textsuperscript{\rm 1},
    Xinyue Liao\textsuperscript{\rm 1},
    Ke-Jia Chen\textsuperscript{\rm 1},
    Songcan Chen\textsuperscript{\rm 2}
}

\affiliations{
    \textsuperscript{\rm 1}School of Computer Science, University of Posts and Telecommunications, Nanjing, China  \\
    \textsuperscript{\rm 2}College of Computer Science and Technology, Nanjing University of Aeronautics and Astronautics, Nanjing, China  \\
    \{wangyunyun, 1223045211, 1224045627, chenkj\}@njupt.edu.cn,
    s.chen@nuaa.edu.cn
}

\usepackage{bibentry}

\definecolor{baseline}{RGB}{235,235,235}
\definecolor{cliprow}{RGB}{226,246,255}
\definecolor{sdrow}{RGB}{238,224,246}
\definecolor{lastrow}{RGB}{215,215,255}
\definecolor{best}{RGB}{255,210,210}

\newcommand{\hl}[1]{\cellcolor{best}{#1}}

\begin{document}

\maketitle

\begin{abstract}
Open-Set Domain Generalization (OSDG) tackles the realistic scenario where deployed models encounter both domain shifts and novel object categories. Despite impressive progress with vision-language models like CLIP, existing methods still fall into the dilemma between structural risk of known-classes and open-space risk from unknown-classes, and easily suffers from over-confidence, especially when distinguishing ``hard unknowns" that share fine-grained visual similarities with known classes. To this end, we propose a Semantic-enhanced CLIP (SeeCLIP) framework that explicitly addresses this dilemma through fine-grained semantic enhancement. In SeeCLIP, we propose a semantic-aware prompt enhancement module to decompose images into discriminative semantic tokens, enabling nuanced vision-language alignment beyond coarse category labels. To position unknown prompts effectively, we introduce duplex contrastive learning with complementary objectives, that is, repulsion to maintain separability from known classes, and cohesion to preserve semantic proximity. Further, our semantic-guided diffusion module synthesizes pseudo-unknowns by perturbing extracted semantic tokens, generating challenging samples that are visually similar to known classes yet exhibit key local differences. These hard negatives force the model to learn finer decision boundaries. Extensive experiments across five benchmarks demonstrate consistent improvements of 3\% accuracy and 5\% H-score over state-of-the-art methods.
\end{abstract}

 \begin{links}
     \link{Appendix}{https://github.com/Leagelab/seeclip}
 \end{links}

\section{Introduction}

Domain Generalization (DG) focuses on training models maintaining robustness when deployed in unseen target domains \cite{wang2022generalizing}. The target domains typically have data distributions that vary substantially from those of the source domains. Traditional DG assumes that all domains share the same label space \cite{li2017deeper,zhou2020learning}, while practical applications often encounter scenarios where the target domain contains not only familiar classes but also entirely new ones. It gives rise to Open-Set Domain Generalization (OSDG) that needs to handle both known and unknown classes under distribution shift \cite{wang2023generalizable}. It is common across various applications \cite{sun2023survey,yang2024open}, such as autonomous driving, remote sensing, and medical imaging, in which detecting unforeseen objects is often critical.

Recently, a few researches have started focusing on OSDG, and the core challenge lies in identifying unknown samples never encountered during training \cite{vaze2022gcd}. Existing methods commonly employ meta-learning to boost classifier learning on known classes \cite{shu2021open,wang2023generalizable}, enabling the model to distinguish unknown classes from known ones during inference. Further, pseudo open-set samples are generated in learning, recasting the open-set recognition as a closed-set problem \cite{singha2024unknown,gupta2025osloprompt}. However, they often struggle to accommodate the variability and complexity of real unknown data. Recently, large-scale vision-language models \cite{radford2021learning,jia2021scaling,cherti2023reproducible} that leverage semantic information in text prompts to better understand images have demonstrated impressive performance for OSDG.

Nevertheless, previous methods have two main limitations. First, they fail to sufficiently model the fine-grained semantics \cite{lang2024coarse}, leading to over-confidence in distinguishing semantically similar categories. That is, they fail to capture the subtle semantic differences between similar categories like ``persian'' and ``siamese" cats. Consequently, they may misclassify unknown objects that have subtle semantic differences from known classes, sacrificing the open-space risk for structural risk. Second, in pseudo-open generation, capturing the real data distribution is often challenging due to its openness and diversity. Generating unknown samples too far from or too close to known categories both leads to model bias \cite{bele2024learning,gupta2025osloprompt}, increasing the generalization risk. In fact, they commonly generate samples that diverge excessively from known classes, e.g., ``desk" vs. known ``cat", causing the model to overlook challenging unknown samples like ``tiger". Consequently, the model becomes biased to known classes and falls into the dilemma between structural risk and open-space risk, leading to sub-optimal performance.

To this end, we propose a Semantic-enhanced CLIP (SeeCLIP) framework that leverages fine-grained semantics in both prompt learning \cite{zhou2022conditional,zhou2022learning} and pseudo-open generation, so as to enable precise discrimination among categories. Specifically, we introduce a semantic-aware prompt enhancement module, which utilizes spatial attention analysis to extract key semantic features from CLIP, and dynamically integrates them into text prompts using learnable weights, establishing a fine-grained vision-language alignment. Duplex contrastive learning is proposed for prompt learning with two complementary constraints. The repulsion loss aims to push the unknown prompts away from known classes, while the cohesion loss restricts their relative distance for semantic proximity. In this way, the unknown prompt will be similar to known prompts, while exhibiting key semantic differences. Moreover, we present a semantic-guided diffusion module, in which the key semantic features extracted are deliberately perturbed and injected into a pre-trained diffusion model as control conditions. Consequently, pseudo-open samples that are globally similar yet locally different from known classes are generated. In fact, these generated samples can be regarded as hard unknowns, which help capture the core semantic features for individual categories, and construct a compact feature space for known classes. In this way, the model considers both structural and open-space risks, and the robustness of open-set recognition will be enhanced.

We also formulate a generalization bound for OSDG, and show that SeeCLIP achieves a lower generalization risk. There are also fine-grained DG researches \cite{yu2024fine,bi2025learning} recently, aiming to identify fine-grained categories within a closed-set environment, and the training data commonly includes structured fine-grained labels. Unlike previous studies, we focus on hard open-set classes without supervised fine-grained knowledge, a significant challenge for OSDG in real-world scenarios. We aim to balance the structural risk of known-classes and open-space risk of unknown-classes, achieving more robust generalization capability. The main contributions are summarized as follows,

\begin{itemize}
    \item We propose SeeCLIP for OSDG, which utilizes fine-grained semantics in both prompt learning and pseudo-open generation, so as to address the under-utilization of fine-grained semantic details, and improve model capability for recognizing indistinguishable unknowns.
    \item We propose a generalization bound for OSDG, under which we show that SeeCLIP accommodates both the structural risk and the open-space risk, achieving a lower generalization risk for OSDG.
    \item Extensive experiments on multiple benchmark datasets demonstrate that SeeCLIP significantly enhances the robustness of open-set recognition. It achieves an average improvement of 3\% in accuracy and approximately 5\% increase in H-score, over the state-of-the-art methods.
\end{itemize}

\section{Related Works}

\subsection{Open-Set Domain Generalization}

OSDG extends traditional DG by requiring models to handle both known and unknown classes under distribution shifts, distinct from both Open Set Recognition (OSR) \cite{bendale2016towards,kong2021opengan} and Open Set Domain Adaptation (OSDA) \cite{panareda2017open}. Pioneering work in OSDG was led by \citeauthor{shu2021open}, who propose a domain-augmented meta-learning framework to simulate open-set scenarios. Subsequent advancements have explored diverse technical paths. Methods like MEDIC \cite{wang2023generalizable} match gradients at both domain and class levels. Adversarial approaches \cite{bose2023beyond,rakshit2022osda} generate pseudo-unknowns for training, recasting open-set recognition as a closed-set problem. Recent efforts have shifted toward integrating prompt tuning. A representative example is ODG-CLIP \cite{singha2024unknown}, which leverages an explicit ``unknown" prompt and a pre-trained diffusion model \cite{rombach2022high} to generate pseudo-open samples, to mimic domain-specific styles while exhibiting semantic divergence. However, previous methods often fail to model fine-grained semantic distinctions or generate pseudo-samples with excessive divergence from known classes, leading to over-confidence boundaries that misclassify unknowns with localized variations.

\subsection{Vision-Language Models and Prompt Learning}

Vision-Language Models (VLMs) like CLIP \cite{radford2021learning} and ALIGN \cite{jia2021scaling} have revolutionized zero-shot recognition through contrastive learning of image-text alignment. Early VLMs relied on hand-crafted prompts, e.g., ``a photo of a \{class\}'', but static prompts lack adaptability to domain-specific nuances or fine-grained semantics. Subsequent prompt learning techniques dynamically optimized prompt tokens for task alignment. CoOp \cite{zhou2022learning} learns continuous prompt embeddings via gradient descent. DualPrompt \cite{wang2022dualprompt} decouples domain-invariant and task-specific prompts to enable multi-domain adaptation. MaPLe \cite{khattak2023maple} further enhances visual-semantic alignment by joint cross-modal prompt tuning. However, they still face challenges in open-set scenarios: 1) They prioritize global feature alignment but overlook localized semantic cues critical for fine-grained discrimination. 2) Hand-crafted or rigidly learned prompts struggle to reject unknowns highly similar to known classes. Thereby, we propose SeeCLIP by incorporating fine-grained semantic cues to effectively reject hard unknown samples, thereby enhancing open-set recognition capability in OSDG.

\section{Proposed Methodology}

\subsection{Task Definition and Model Architecture}

\begin{figure*}[t]
\centering
\includegraphics[width=1\textwidth]{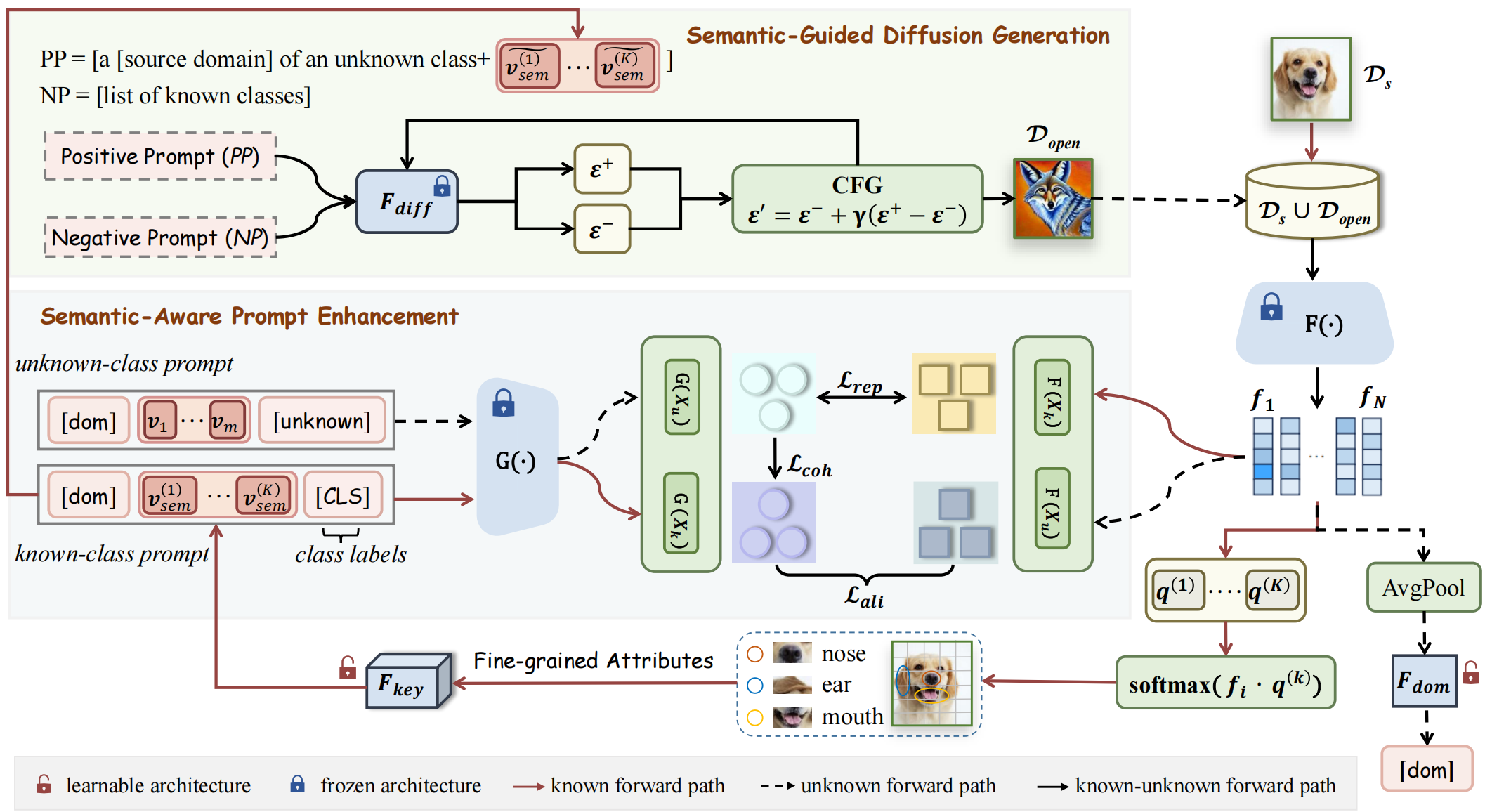}
\caption{Overall architecture of SeeCLIP, which consists of a semantic-aware prompt enhancement module for boosting prompt learning with fine-grained semantics, and a semantic-guided diffusion module to generate pseudo-unknowns as hard unknowns.}
\label{fig1}
\end{figure*}

In OSDG, we consider multiple source domains, each with its own distinct data distribution, yet sharing a common set of classes. Formally, given $M$ source domains \{$\mathcal{D}_1, \dots, \mathcal{D}_M$\}, where each $\mathcal{D}_k = \{ (x_i^s, y_i^s) \}_{i=1}^{n_k}$ consists of $n_k$ labeled samples from $C$ categories. The target domain $\mathcal{D}_t = \{ x_j^{t} \}_{j=1}^{n_t}$ exhibits a different distribution from source domain. It contains unlabeled samples that may belong to known classes or novel classes not encountered during training. The known label space is $\mathcal{Y}^s = \{1, \ldots, C\}$, and the unknown label space in the target domain is $\mathcal{Y}^u = \{C+1\}, \quad \mathcal{Y}^t = \mathcal{Y}^s \cup \mathcal{Y}^u$. The learning goal is to develop a robust classifier capable of distinguishing between known and unknown classes in the target domain, while classifying samples from known classes accurately.

We propose a SeeCLIP framework, which enhances OSDG by effectively leveraging fine-grained semantics, so as to boost the robustness of open-set recognition. SeeCLIP consists of a semantic-aware prompt enhancement module, and a semantic-guided diffusion module, as shown in Figure~\ref{fig1}. In semantic‑aware prompt enhancement learning, fine‑grained semantic tokens $\left\{ v_{\text{sem}}^{(1)}, \ldots, v_{\text{sem}}^{(K)} \right\}$ are extracted via $K$-head attention using learnable queries $\{ q^{(1)}, \ldots, q^{(K)}\}$, and combined with domain tokens [dom] to construct enhanced prompts. For unknown prompts, learnable class-agnostic semantic tokens $\{v_{1}, \ldots, v_m\}$ and the token [$unknown$] are used. In semantic-guided diffusion module, perturbed semantic tokens are fused with a positive prompt $PP$ to form the positive condition, while a negative prompt $NP$ listing all known classes serves as a counter-guidance to generate pseudo-unknowns. Classifier-Free Guidance (CFG)\cite{ho2022classifier} is employed to strengthen the distinction between positive and negative conditions. Three losses, including alignment, repulsion, and cohesive losses, jointly optimize the prompts, enforcing a clear boundary between known and unknown categories.

\subsection{Semantic-Aware Prompt Enhancement}

To distinguish semantically similar unknowns in open domains, we aim to enhance prompt learning with both domain-level semantics and multi-token discriminative features, enabling fine-grained vision-language alignment under distribution shifts.

\textbf{\textit{Representation of Prompts.}} The input data is encoded by the vision encoder of CLIP, then we build a domain token $v^{(k)}_{dom}$ for source domain $\mathcal{D}_k$ by a simple average over all samples. Each sample is also transformed into a sequence of patch embeddings ${\{f_i\}}^N_{i=1}$, where each ${f_i} \in \mathrm{R}^d$ represents the feature embedding of a localized region. To extract fine-grained semantics, we adopt a $K$-head attention pooling mechanism,considering that only a few visual heads in attention mechanisms dominate visual understanding by focusing on key image regions linked to fine-grained semantics\cite{zhang2025less}. $K$ learnable query vectors ${\{q^{(k)}\}}^K_{k=1}$ are introduced for each class, each focusing on a specific semantic region, and are shared across all domains. An attention weight $w^{(k)}_i$ is assigned for the $k$-th attention head,
\begin{equation}
\omega_i^{(k)} = \frac{\exp \left( q^{(k)} \cdot f_i \right)}{\sum_{j=1}^{N} \exp \left( q^{(k)} \cdot f_j \right)}
\end{equation}

The $k$-th semantic token $v_{sem}^{(k)}$ is then formulated as a weighted aggregation of all patch embeddings, i.e.,
\begin{equation}
v_{sem}^{(k)} = \sum_{i=1}^{N} \omega_i^{(k)} \cdot f_i
\end{equation}
It generates $K$ semantic tokens, each adept at capturing the discriminative features from a distinct part, e.g., tail, eyes, ears.
Further, to construct an enhanced prompt for each known class, we integrate both domain and semantic tokens using learnable projections $\Phi(\cdot)$ and $\Psi(\cdot)$, respectively,
\begin{equation}
p_c = [\Phi(v_{\text{dom}})], [\Psi_1(v_{\text{sem}}^{(1)}), \ldots, \Psi_K(v_{\text{sem}}^{(K)})], [\textit{classname}]
\end{equation}

For unknown categories lacking semantic description, we construct the prompt as,
\begin{equation}
p_{{unk}} = [\Phi(v_{\text{dom}})], [v_1, \ldots, v_m], [\textit{unknown}]
\end{equation}
where tokens $v_1,\ldots,v_m$ are a set of learnable semantic vectors encoding generalizable, class-agnostic patterns frequently observed in unknown classes. $[unknown]$ serves as a trainable label token, indicating that the prompt corresponds to an unknown category. It helps represent and distinguish samples that fall outside known-class distribution. 

\textbf{\textit{Learning for Class Prompts.}} We introduce a duplex contrastive formulation for prompt learning. To enforce a clear open-set boundary, a margin-based repulsion loss pushes the unknown prompt away from all known embeddings,
\begin{equation}
\mathcal{L}_{{rep}} =
        \sum_{c=1}^{C} \max(0, \delta - \text{sim}(F_{t}(p_{\text{unk}}), F_{v}(\mathbf{X}_c)))
\end{equation}
where $\delta$ is the margin hyperparameter, $F_t$ and $F_v$ are the text and vision encoders of CLIP, respectively, $\mathbf{X}_c$ denotes data from the $c$-th class. Besides, to prevent unknown prompt from drifting too far from known categories, we encourage its proximity to the center of known prompt embeddings with a cohesive loss,
\begin{equation}
\mathcal{L}_{{coh}}=
        \left\| F_{t}(p_{\text{unk}}) - \frac{1}{C} \sum_{c=1}^{C} F_{t}(p_{c}) \right\|_{2}^{2}
\end{equation}

We also apply an $L_1$ regularization to each projected semantic token to suppress overfitting and encourage a sparse utilization of fine-grained semantics, i.e.,
\begin{equation}
\mathcal{L}_{{reg}} = \lambda \sum_{k=1}^{K} \left\| \Psi_k \left( v_{\text{sem}}^{(k)} \right) \right\|_1
\end{equation}
In this way, the model can automatically select the most discriminative semantic components while filtering out noisy or redundant ones, bringing more robust and transferable representations for open-set recognition across different domains.

\subsection{Semantic-Guided Diffusion Generation}

Building on the semantic token representations, we aim to synthesize pseudo-unknown samples that globally resemble known classes while exhibiting local deviations, focusing on the core semantics of known categories.

We first perturb each semantic token with Gaussian noise,
\begin{equation}
\widetilde{v_{\text{sem}}^{(k)}} = v_{sem}^{(k)} + \epsilon_k, \quad \epsilon_k \sim \mathcal{N}(0, \sigma^2 I)
\end{equation}
where $\sigma$ regulates the perturbation intensity. A small $\sigma$ ensures coherence semantic with known classes, yet discriminative local variations. The perturbed tokens serve as conditional inputs to the diffusion model.

We use a positive textual prompt describing the source domain with unknown-class context,
\begin{center}
{\spaceskip=0.3em plus 0.1em minus 0.1em
$PP$ = $\text{``A [\textit{source domain}] image of an unknown class"}$}
\end{center}
where [\textit{source domain}] is the specific domain name, such as "photo" or "art painting" in PACS dataset. Its embedding $E_{text}^+$ is then concatenated with visual condition 
$E_{\text{visual}} = \left[ \Psi_1\left( \widetilde{v_{\text{sem}}^{(1)}} \right), \ldots, \Psi_K\left( \widetilde{v_{\text{sem}}^{(K)}} \right) \right]$ to form a joint condition,
\begin{equation}
E_{\text{joint}} = \left[ E_{\text{text}}^+; E_{\text{visual}} \right]
\end{equation}

It is subsequently fed into the cross-attention layers of the diffusion network, ensuring that the generated samples preserve global domain structure while introducing controlled visual divergence.

Further, to suppress the high belongingness to known categories, we introduce a negative textual prompt including all known-class names,
\begin{center}
{\spaceskip=0.3em plus 0.1em minus 0.1em
 $NP$ = $\text{``[\textit{Known Class} 1], ..., [\textit{Known Class C}] "}$}
\end{center}

By integrating positive and negative prompt conditions, model will generate pseudo unknowns that balance semantic validity with distributional novelty. It ensures that the generated samples mimic real-world open-set categories, thereby enhancing generalization to unseen classes.

\subsection{Loss Functions, Training, and Inference}

SeeCLIP learns by simultaneously aligning image representations with their corresponding prompts, enforcing separation between known and unknown prompts, and regularizing semantic token projections to prevent overfitting as well. 

A symmetric contrastive alignment loss is adopted to pull each sample and the corresponding prompt closer in the shared embedding space,
\begin{equation}
\mathcal{L}_{{ali}} =
        -\log \frac{\exp(\text{sim}(F_{v}(x_{i}), F_{t}(p_{i})) / \tau)}
       {\sum\nolimits_{j=1}^{C} \exp(\text{sim}(F_{v}(x_{i}), F_{t}(p_{j})) / \tau}
\end{equation}
where $p_i$ is the class prompt of $x_i$, $p_j$ is the prompt of classes that $x_i$ does not belong to, $\tau$ is a temperature scaling factor. 

Finally, the total loss for SeeCLIP is defined as,
\begin{equation}
\mathcal{L} = \mathcal{L}_{{ali}} + \alpha \mathcal{L}_{{rep}} + \beta \mathcal{L}_{{coh}} + \gamma \mathcal{L}_{{reg}}
\end{equation}
where $\alpha$, $\beta$ and $\gamma$ are hyper-parameters.

During training, the visual and textual encoders of CLIP are frozen to retain pre-trained semantic stability. Optimization is performed only on the learnable query vectors, the semantic/domain token projection layers, and the unknown prompt embeddings. After the diffusion generation, the end-to-end training procedure utilizes both known-class and synthesized pseudo-unknown samples, alternating between two phases: i) the alignment phase, which aligns known visual embeddings with enhanced prompts, and ii) the repulsion phase, which pushes the unknown prompt away from known features while preventing excessive semantic drift. In inference, samples are predicted by similarity with enhanced prompts, including both known and unknown class prompts, as shown in Algorithm~\ref{alg:seec}.

\begin{algorithm}[t]
\caption{Training of SeeCLIP}
\label{alg:seec}
\begin{algorithmic}[1]
\REQUIRE Source domains $\{\mathcal{D}_{k}\}_{k=1}^{M}$, frozen CLIP encoders
\ENSURE Semantic tokens and projections

\STATE Initialize $\{q^{(k)}\}_{k=1}^{K}$, $\Phi(\cdot)$, $\{\Psi_{k}(\cdot)\}_{k=1}^{K}$, and $\{v_{i}\}_{i=1}^{m}$

\STATE \textbf{for} each training iteration \textbf{do}
    \STATE \hspace{1em} \textbf{for} each domain $\mathcal{D}_k$ \textbf{do}
        \STATE \hspace{2em} $v_{\text{dom}}^{(k)} \leftarrow \frac{1}{|F_{k}|}\sum_{f_i \in F_k}f_i $
        \STATE \hspace{2em} \textbf{for} each attention head $k$ \textbf{do}
            \STATE \hspace{3em} $\omega_i^{(k)} \leftarrow \frac{\exp(q^{(k)} \cdot f_i)}{\sum_{j=1}^{N} \exp(q^{(k)} \cdot f_j)}$ as in Eq. (1)
            \STATE \hspace{3em} $v_{\text{sem}}^{(k)} \leftarrow \sum_{i=1}^{N} \omega_i^{(k)} \cdot f_i$ as in Eq. (2)
        \STATE \hspace{2em} \textbf{end for}
    \STATE \hspace{1em} \textbf{end for}
     \STATE \hspace{1em} $p_c \leftarrow [\Phi(v_{\text{dom}})], [\Psi_1(v_{\text{sem}}^{(1)}), \ldots, \Psi_K(v_{\text{sem}}^{(K)})], [\textit{classname}]$
    \STATE \hspace{1em} $p_{\text{unk}} \leftarrow [\Phi(v_{\text{dom}})], [v_1, \ldots, v_m], [\textit{unknown}]$
    \STATE \hspace{1em} $\widetilde{v_{\text{sem}}^{(k)}} \gets v_{\text{sem}}^{(k)} + \epsilon_k$, $\epsilon_k \sim \mathcal{N}(0, \sigma^2 I)$ as in Eq. (8)
    \STATE \hspace{1em} Generate pseudo-unknowns via diffusion with $[\widetilde{v_{\text{sem}}^{(k)}}]$
    \STATE \hspace{1em} $\mathcal{L} \leftarrow \mathcal{L}_{{ali}} + \alpha \mathcal{L}_{{rep}} + \beta \mathcal{L}_{{coh}} + \gamma \mathcal{L}_{{reg}}$ as in Eq. (11)
    \STATE \hspace{1em} Update parameters via gradient descent

\STATE \textbf{end for} 
\end{algorithmic}
\end{algorithm}

\subsection{Theoretical Analysis}

\textbf{Theorem 1 (OSDG error bound)} Let ${\gamma} := \min_{\pi \in \Delta_M} d_\mathcal{H} \left( \mathcal{P}_\mathcal{X}^t, \sum_{i=1}^M \pi_i \mathcal{P}_\mathcal{X}^i \right)$ with minimizer $\pi^*$ be the distance of $\mathcal{P}_\mathcal{X}^t$ from the convex hull of sources $\Lambda$, and $\mathcal{P}_\mathcal{X}^* := \sum_{i=1}^M \pi_i^* \mathcal{P}_\mathcal{X}^i$ be the best approximator within $\Lambda$. $\rho :=  \sup_{{\mathcal{P}_{\mathcal{X}'}}, {\mathcal{P}_{\mathcal{X}''}} \in \Lambda} d_H({\mathcal{P}_{\mathcal{X}'}}, \mathcal{P}_{\mathcal{X}''})$ is the diameter of $\Lambda$. $\pi^{unk} = Pr(\mathcal{Y}^t \in \mathcal{Y}^{u})$ denotes the prior probabilities of unknown class in the target domain. Then for any hypothesis $h \in \mathcal{H}$, the target risk $\mathcal{R}^t(h)$ is,

\begin{equation}
\mathcal{R}^t(h) \leq \sum_{i=1}^{M} \pi_i^* \mathcal{R}^i(h) + \frac{\gamma + \rho}{2} + \lambda_{\mathcal{H}(P_\mathcal{X}^t, P_\mathcal{X}^*)} + \pi^{unk} \cdot \mathcal{R}^{\text{OS}}(h)
\end{equation}
where $\mathcal{R}^i(h)$ is the risk of the $i$-th source domain. $\lambda_{\mathcal{H}(\mathcal{P}_\mathcal{X}^t, \mathcal{P}_\mathcal{X}^*)}$ is the ideal joint risk across the target domain and the domain with the best approximator distribution $\mathcal{P}_\mathcal{X}^*$. $\mathcal{R}^{\text{OS}}(h) = \mathbf{E}_{(x,y) \sim \mathcal{D}^t \cap \mathcal{Y}^u} \mathbf{I}(h(x) \in \mathcal{Y}^s)$ is the open-space risk, which represents the risk of misclassifying unknown-class samples in target domain to known classes.

\noindent\textbf{{Lemma 1}} Let $dis(c,d)$ and $dis_{\text{sem}}{(c,d)}$ denote the discrepancy between original and enhanced prompts of the $c$-th and $d$-th classes, respectively, $\forall c,d \in \{1,...,C\}$, $c\neq d$, then it holds that ${dis_{\text {sem}}(c,d)>dis(c,d)}$.

\noindent{\textbf{Remark 1.}} From Lemma 1, by introducing fine-grained semantics, discrepancy between class prompts will be enlarged. Consequently, the inter-class discrimination will be enhanced, which helps reduce the weighted source structural risks $\sum_{i=1}^{M} \pi_i^* \mathcal{R}^i(h)$ in $\mathcal{R}^t(h)$.

\noindent\textbf{{Remark 2.}} The high-similarity pseudo-unknowns force model to learn core features of known classes, reducing its over-generalization to unknown classes. They also help compress the feature space of known classes, thereby help lower the open-space risk $\mathcal{R}^{\text{OS}}(h)$.
The proof can be found in the Appendix. SeeCLIP addresses both risks, thus effectively lowers the generalization risk of OSDG.

\section{Experiments}

\subsection{Experimental Setups}

\noindent\textbf{Datasets.}
We evaluate SeeCLIP on five benchmark datasets: Office-Home, PACS, VLCS, Mini-DomainNet, and Multi-Dataset, following standard known-novel class splits. Office-Home contains 65 categories across four domains. PACS comprises 7 categories with four domains. VLCS includes 5 categories across four domains. Mini-DomainNet contains 126 categories with four domains. Multi-Dataset combines samples from Office-Home, PACS, and VLCS. For each dataset, we follow the leave-one-domain-out protocol, where one domain is selected as the target and the remaining domains are used as sources for training.

\noindent\textbf{Evaluation Metrics.}
Following standard protocols in OSDG, we evaluate with two key metrics: 1) top-1 accuracy (ACC) for closed-set classes, and 2) H-score for unknown class detection, which represents the harmonic mean of known and unknown accuracies.

\noindent\textbf{Compared Methods.} 
We compare SeeCLIP with two broad families of baselines: 
1) Traditional OSDG/OSR methods built on ResNet-18, with a confidence-threshold rule to reject unknowns: Cumix, MixStyle, DAML, and MEDIC. 
2) CLIP-based models, evaluated under the same leave-one-domain-out protocol. 
We also cite three strong closed-set DG methods for context: SWAD, EoA, and DandelionNet.

\noindent\textbf{Architecture Details.}
For all experiments, we implement SeeCLIP on the CLIP ViT-B/32 architecture, with Transformer serving as the textual encoder. During training, we freeze the parameters of both vision and text encoders, and optimize only the learnable components specific to SeeCLIP. The number of semantic token heads $K$ is set to 4, and the prompt tokens number $M$ for unknown class is set to 3. We employ the Stable Diffusion v1.5 with 50 denoising steps and a perturbation standard deviation $\sigma$ of 0.2.

\noindent\textbf{Training and Evaluation.}
We train for 10 epochs using the AdamW optimizer with a learning rate of 1e-4. Batch sizes are set per dataset, i.e., 6 for PACS/VLCS, 9 for Office-Home, Multi-Dataset, and Mini-DomainNet, with each batch incorporating three pseudo-open samples from each source domain. The textual prompt context length is set to 4. We set $\alpha = 0.5, \beta = 0.3, \text{ and } \gamma = 0.1$. The margin $\delta$ is set to 0.2, and $\tau$ is set to 0.07.

\subsection{Performance Comparison}

\subsubsection{OSDG Setting.}

Table~\ref{tab:comparative_results} presents comparison results across five benchmark datasets under the OSDG setting.

\begin{table*}[t]
\caption{Comparative results under the OSDG setting over leave-one-domain-out combinations. \textit{SD} denotes \textit{stable diffusion}.}
\label{tab:comparative_results}
\centering
\small
\setlength{\tabcolsep}{3.5pt}
\begin{tabularx}{\textwidth}{@{}l *{10}{>{\centering\arraybackslash}X} | *{2}{>{\centering\arraybackslash}X}@{}}
\toprule
\multirow{2}{*}{Methods} & 
\multicolumn{2}{c}{PACS} & 
\multicolumn{2}{c}{VLCS} & 
\multicolumn{2}{c}{Office Home} & 
\multicolumn{2}{c}{Multi-Dataset} & 
\multicolumn{2}{c}{Mini-DomainNet} & 
\multicolumn{2}{c@{}}{Average} \\
\cmidrule(r){2-3} \cmidrule(r){4-5} \cmidrule(r){6-7} \cmidrule(r){8-9} \cmidrule(r){10-11} \cmidrule(r){12-13}
 & Acc & H-score & Acc & H-score & Acc & H-score & Acc & H-score & Acc & H-score & Acc & H-score \\
\midrule
\textit{CNN-based} \\
Cumix & 57.85 & 41.05 & 52.46 & 50.11 & 51.67 & 49.40 & 42.18 & 46.91 & 50.27 & 39.16 & 52.09 & 46.81 \\
MixStyle & 63.35 & 48.30 & 52.30 & 50.61 & 53.52 & 49.53 & 42.18 & 46.91 & 50.43 & 40.25 & 53.67 & 48.66 \\
DAML & 65.49 & 51.88 & 53.53 & 51.59 & 56.45 & 53.34 & 46.61 & 51.71 & 52.81 & 43.63 & 55.73 & 51.29 \\
MEDIC & 89.81 & 83.03 & 57.28 & 55.73 & 60.26 & 57.91 & 50.74 & 53.13 & 55.29 & 45.71 & 66.11 & 60.30 \\
\midrule
\textit{CLIP-based} \\
\cellcolor{baseline} CLIP & 95.16 & 76.77 & 91.84 & 72.94 & 81.43 & 63.62 & 77.88 & 72.19 & 84.50 & 68.94 & 84.65 & 69.40 \\
\cellcolor{baseline} CLIP+OpenMax & 93.45 & 79.13 & 92.09 & 73.67 & 81.00 & 61.54 & 78.34 & 73.26 & 81.89 & 69.40 & 83.95 & 69.96 \\
\cellcolor{baseline} CLIP+OSDA & 92.62 & 75.40 & 90.21 & 70.89 & 82.58 & 67.35 & 74.45 & 75.22 & 82.00 & 73.62 & 83.73 & 71.36 \\
\cellcolor{cliprow} CoOp & 78.77 & 26.87 & 92.02 & 39.26 & 73.85 & 36.26 & 66.03 & 44.34 & 61.13 & 68.34 & 71.72 & 41.65 \\
\cellcolor{cliprow} CoCoOp & 85.76 & 32.93 & 89.47 & 37.01 & 75.38 & 34.38 & 64.84 & 47.57 & 60.63 & 56.30 & 71.48 & 40.28 \\
\cellcolor{cliprow} MaPLe & 93.97 & 48.47 & 89.70 & 43.33 & 79.47 & 33.06 & 69.34 & 62.20 & 74.67 & 60.57 & 79.62 & 48.58 \\
\cellcolor{cliprow} LASP & 88.45 & 30.37 & 90.67 & 39.41 & 76.13 & 34.52 & 66.78 & 50.22 & 62.34 & 61.56 & 74.21 & 41.89 \\
\cellcolor{cliprow} PromptSRC & 94.53 & 43.32 & 90.13 & 42.78 & 80.21 & 36.40 & 65.51 & 59.45 & 73.60 & 62.56 & 79.89 & 48.13 \\
\cellcolor{cliprow} CLIPN & 96.24 & 45.00 & 84.82 & 50.72 & 84.55 & 42.83 & 77.16 & 62.60 & 77.38 & 66.92 & 83.64 & 52.27 \\
\cellcolor{cliprow} STyLIP & 95.36 & 50.74 & 90.75 & 65.66 & 84.73 & 60.97 & 79.88 & 71.99 & 80.22 & 69.11 & 85.26 & 62.77 \\
\cellcolor{cliprow} CLIPN+STyLIP & 96.37 & 64.46 & 84.65 & 68.02 & 83.67 & 76.50 & 76.93 & 72.15 & 86.59 & 76.18 & 85.06 & 69.43 \\
\cellcolor{sdrow} MaPLe+SD & 91.47 & 82.60 & 91.70 & 72.67 & 85.02 & 80.60 & 77.62 & 72.83 & 83.79 & 79.30 & 84.92 & 75.64 \\
\cellcolor{sdrow} PromptSRC+SD & 93.21 & 87.95 & 90.34 & 72.62 & 84.60 & 83.31 & 78.44 & 77.89 & 83.87 & 82.95 & 85.23 & 78.35 \\
\cellcolor{sdrow} STyLIP+SD & 91.78 & 87.42 & 92.11 & 73.34 & 85.51 & 81.22 & 79.05 & 78.52 & 84.12 & 83.21 & 85.67 & 78.64 \\
\cellcolor{sdrow} ODG-CLIP & 
\hl{99.53} & \hl{99.70} &
\hl{95.71} & \hl{86.53} &
\hl{98.32} & \hl{96.08} &
\hl{84.60} & \hl{90.00} &
\hl{95.68} & \hl{94.48} &
\hl{94.23} & \hl{90.84} \\
\midrule
\rowcolor{lastrow}
\textbf{SeeCLIP} & 99.90 & 99.97 & 98.30 & 89.49 & 99.50 & 98.97 & 88.70 & 92.80 & 98.87 & 97.06 & \textbf{97.05} & \textbf{95.66} \\
\bottomrule
\end{tabularx}
\end{table*}

\noindent\textit{Better Performance among Compared Methods.} As shown in Table~\ref{tab:comparative_results}, CNN-based methods show limited effectiveness in OSDG. Standard CLIP-based approaches perform better, but prompt learning methods without explicit unknown handling yield poor H-scores despite reasonable accuracy. SeeCLIP outperforms all baselines across metrics, validating our proposal. It achieves consistent gains across datasets and exhibits robustness to diverse domain shifts. On PACS with artistic style variations, it delivers near-perfect results. Its 0.27\% H-score improvement over ODG-CLIP seems modest yet meaningful given the high baseline. On VLCS, it outperforms ODG-CLIP by 2.59\% and 2.96\% respectively, highlighting superior cross-domain generalization.
\noindent\textit{Effectiveness in Fine-grained Recognition.} SeeCLIP exhibits its most significant gains on Office-Home, outperforming ODG-CLIP by 1.18\% and 2.89\%, respectively. Given the challenges posed by fine-grained object categories across diverse domains for unknown class detection, these improvements are particularly notable. The results validate the our effectiveness in capturing subtle semantic differences, critical for distinguishing similar classes.

\subsubsection{DG Setting.}

We also evaluate SeeCLIP under the standard DG setting. As shown in Table~\ref{tab:standard_dg_results}, SeeCLIP shows exceptional performance across all benchmarks. It significantly outperforms traditional CNN-based methods and maintains substantial performance gains when compared against CLIP-based methods. Most notably, SeeCLIP achieves superior performance compared to the previous state-of-the-art method ODG-CLIP, with improvements of 0.50\% in accuracy, demonstrating its effectiveness in closed-set setting.

\begin{table}[t]
\caption{Performance comparison in standard DG setting.}
\label{tab:standard_dg_results}
\centering
\small
\setlength{\tabcolsep}{4.5pt}
\begin{tabular}{@{}l c c c c c@{}}
\toprule
\textbf{Methods} & 
\textbf{PACS} & 
\textbf{VLCS} & 
\textbf{O. H.} & 
\textbf{M. DNet} & 
\textbf{Avg.} \\
\midrule
{\textit{CNN-based}} \\
SWAD & 88.10 & 79.10 & 70.60 & -- & 79.27 \\
EoA & 88.60 & 79.10 & 72.50 & -- & 80.07 \\
DandelionNet & 89.20 & 81.60 & 70.40 & -- & 80.40 \\
\midrule
{\textit{CLIP-based}} \\

\cellcolor{baseline} CLIP & 94.89 & 82.14 & 78.40 & 78.73 & 83.54 \\
\cellcolor{cliprow} CoOp & 97.11 & 83.34 & 81.33 & 72.30 & 83.52 \\
\cellcolor{cliprow} CoCoOp & 96.54 & 85.02 & 81.05 & 71.51 & 83.53 \\
\cellcolor{cliprow} MaPLe & 97.72 & 86.75 & 83.52 & 73.87 & 85.47 \\
\cellcolor{cliprow} LASP & 97.02 & 87.25 & 84.13 & 70.67 & 84.77 \\
\cellcolor{cliprow} PromptSRC & 98.02 & 86.34 & 83.89 & 76.10 & 86.09 \\
\cellcolor{cliprow} StyLIP & 98.17 & 87.21 & 85.94 & 80.43 & 87.94 \\
\cellcolor{sdrow} ODG-CLIP & \hl{99.83} & \hl{95.74} & \hl{96.91} & \hl{96.65} & \hl{97.28} \\
\rowcolor{lastrow}
SeeCLIP & \textbf{99.89} & \textbf{96.52} & \textbf{97.43} & \textbf{97.28} & \textbf{97.78} \\
\bottomrule
\end{tabular}
\end{table}

\subsection{Ablation Analysis}

\begin{table}[t]
\centering
\caption{Ablation study of individual modules in SeeCLIP.}
\label{tab:ablation_results}
\small
\setlength{\tabcolsep}{4.5pt}
\resizebox{\columnwidth}{!}{
\begin{tabular}{@{}lcc cc cc cc@{}}
\toprule
\multirow{2}{*}{\makecell{\textbf{Methods}}} & 
\multicolumn{2}{c}{\textbf{PACS}} & 
\multicolumn{2}{c}{\textbf{O. H.}} & 
\multicolumn{2}{c}{\textbf{M. Data}} & 
\multicolumn{2}{c}{\textbf{M. DNet}} \\
\cmidrule(lr){2-3} \cmidrule(lr){4-5} \cmidrule(lr){6-7} \cmidrule(lr){8-9}
& \textbf{Acc} & \textbf{H} & \textbf{Acc} & \textbf{H} & \textbf{Acc} & \textbf{H} & \textbf{Acc} & \textbf{H} \\
\midrule
Baseline & 95.16 & 76.77 & 81.43 & 63.62 & 77.88 & 72.19 & 84.50 & 68.94 \\
+ SAPE & 97.34 & 88.42 & 91.67 & 84.73 & 84.23 & 86.51 & 93.45 & 87.29 \\
+ SGDG & 96.28 & 82.15 & 87.92 & 76.84 & 81.56 & 79.67 & 89.73 & 81.52 \\
+ DCL & 96.89 & 85.67 & 89.45 & 81.92 & 83.12 & 82.34 & 91.67 & 84.78 \\
+ SAPE \& SGDG & 98.45 & 92.78 & 95.23 & 91.45 & 86.78 & 89.67 & 96.34 & 93.12 \\
+ SAPE \& DCL & 98.89 & 94.56 & 96.78 & 93.82 & 87.45 & 90.89 & 97.23 & 94.67 \\
+ SGDG \& DCL & 97.67 & 91.23 & 93.56 & 88.34 & 85.34 & 87.78 & 94.78 & 90.45 \\
\midrule
\textbf{Full (SeeCLIP)} & \textbf{99.90} & \textbf{99.97} & \textbf{99.50} & \textbf{98.97} & \textbf{88.70} & \textbf{92.80} & \textbf{98.87} & \textbf{97.06} \\
\bottomrule
\end{tabular}}
\end{table}

\begin{table}[t]
\centering
\caption{Ablation analysis of loss functions in SeeCLIP.}
\label{tab:ablation_loss}
\small
\setlength{\tabcolsep}{4.5pt}
\resizebox{\columnwidth}{!}{
\begin{tabular}{@{}l *{8}{c} @{}}
\toprule[1.5pt]
\multirow{2}{*}{\textbf{Configuration}} & 
\multicolumn{2}{c}{\textbf{PACS}} & 
\multicolumn{2}{c}{\textbf{O. H.}} & 
\multicolumn{2}{c}{\textbf{M. Data}} & 
\multicolumn{2}{c}{\textbf{M. DNet}} \\
\cmidrule{2-3} \cmidrule{4-5} \cmidrule{6-7} \cmidrule{8-9}
& \textbf{Acc} & \textbf{H} & \textbf{Acc} & \textbf{H} & \textbf{Acc} & \textbf{H} & \textbf{Acc} & \textbf{H} \\
\midrule
\textbf{Full (SeeCLIP)} & \textbf{99.9} & \textbf{99.7} & \textbf{99.5} & \textbf{99.0} & \textbf{88.7} & \textbf{92.8} & \textbf{98.9} & \textbf{97.1} \\
w/o $\mathcal{L}_{\text{ali}}$ & 97.8 & 97.3 & 97.5 & 96.3 & 85.0 & 89.0 & 95.1 & 93.2 \\
w/o $\mathcal{L}_{\text{rep}}$ & 98.6 & 98.0 & 98.3 & 97.4 & 86.9 & 91.5 & 97.0 & 95.5 \\
w/o $\mathcal{L}_{\text{coh}}$ & 99.1 & 98.9 & 99.0 & 98.7 & 88.0 & 92.0 & 98.3 & 96.6 \\
w/o $\mathcal{L}_{\text{reg}}$ & 99.5 & 99.3 & 99.3 & 98.9 & 88.4 & 92.6 & 98.7 & 96.9 \\
\bottomrule[1.5pt]
\end{tabular}}
\end{table}

\noindent\textit{Analysis for Components.} 
The individual components in SeeCLIP reveal distinct contributions relative to the baseline, as shown in Table~\ref{tab:ablation_results}. Semantic-Aware Prompt Enhancement(SAPE) drives the largest improvements, boosting accuracy by 6.93\% and H-score by 16.36\%, validating that fine-grained semantic tokens enhance the discrimination of similar categories. Semantic-Guided Diffusion Generation(SGDG) also yields meaningful gains, increasing accuracy by 4.13\% and H-score by 9.67\%, by generating samples that help sharpen the decision boundaries. Meanwhile, Duplex Contrastive Learning(DCL) achieves notable improvements, with accuracy up 5.54\% and H-score up 13.30\% through effective feature space separation.

\noindent\textit{Analysis for Losses.} Table~\ref{tab:ablation_loss} summarizes the ablation results for each loss component. The alignment loss contributes most significantly: its removal reduces accuracy by 2.89\% and H-score by 3.15\%, confirming its critical role. The repulsion loss is also essential, with its exclusion decreasing accuracy by 1.54\% and H-score by 1.60\%. The cohesion loss, though individually modest, is crucial for semantic coherence. Its removal lowers accuracy by 0.64\% and H-score by 0.65\%. Besides, the regularization loss also provides consistent improvement.

\subsection{Image Generation Analysis}

\begin{figure}[t]
\centering
\includegraphics[width=0.8\columnwidth]{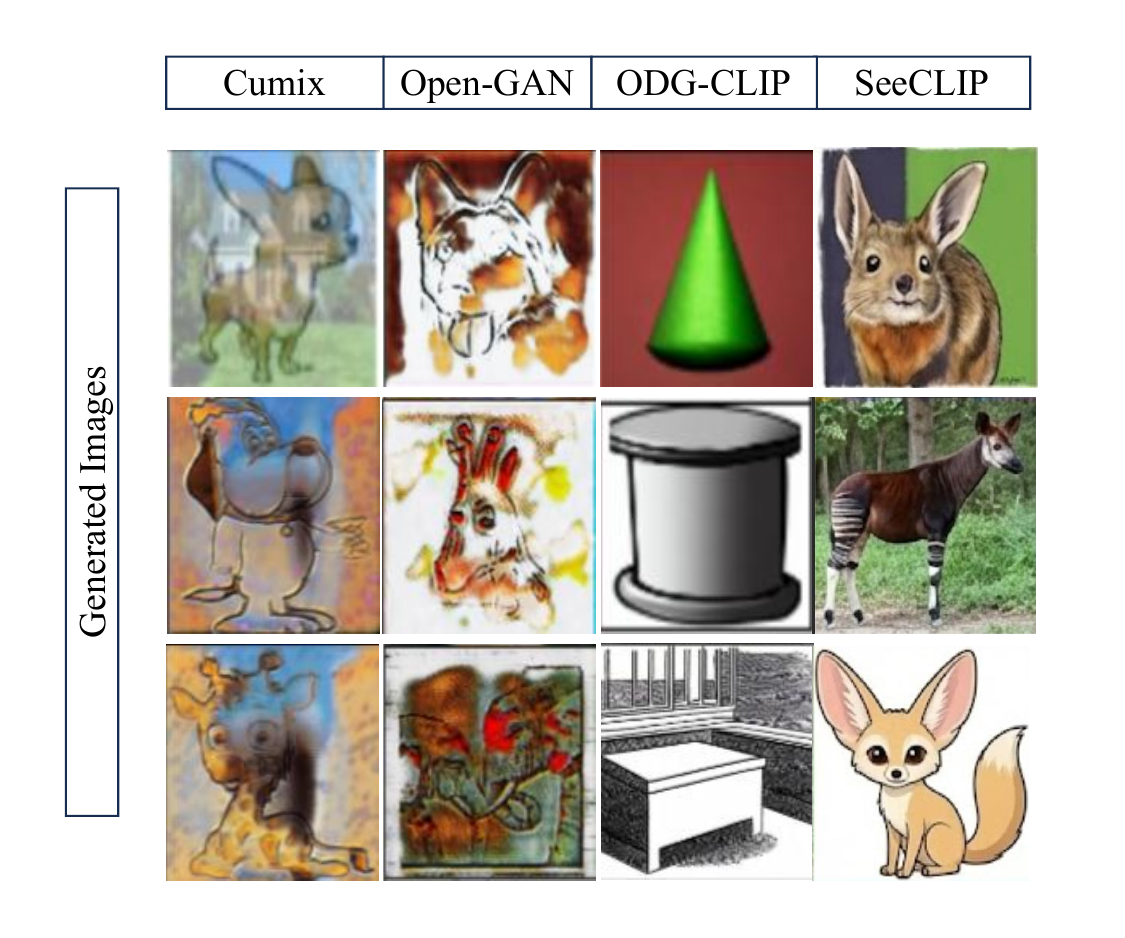}
\caption{Comparison of unknown sample generation methods w.r.t. known classes over PACS. SeeCLIP generates semantically coherent pseudo-unknowns that are globally similar yet locally distinctive from known classes.}
\label{fig:2}
\end{figure}

We compare the unknown generation of SeeCLIP against three established methods in Table~\ref{tab:pseudo_open_generation}, from which SeeCLIP with SGDG achieves the best performance. In Figure~\ref{fig:2}, both Cumix and Open-GAN generate semantically ambiguous pseudo-open images, ODG-CLIP generates unknowns far from the known samples, thus may lead to overfitting to known classes. SeeCLIP adopts semantic token perturbation with Gaussian noise and dual-prompt conditioning to generate pseudo-unknowns. The pseudo-unknowns from SeeCLIP are similar to known classes, yet exhibit key semantic discrepancies. It helps generate a clear boundary that accounts for both risks of known and unknown categories.

\begin{table}[h]
\centering
\caption{Comparison of pseudo-open sample generation.}
\label{tab:pseudo_open_generation}
\small
\setlength{\tabcolsep}{4.5pt}
\resizebox{\columnwidth}{!}{
\begin{tabular}{@{}lcc cc cc cc@{}}
\toprule
\multirow{2}{*}{\textbf{SeeCLIP with}} & 
\multicolumn{2}{c}{\textbf{PACS}} & 
\multicolumn{2}{c}{\textbf{O.H.}} & 
\multicolumn{2}{c}{\textbf{M.Data}} & 
\multicolumn{2}{c}{\textbf{M.DNet}} \\
\cmidrule(lr){2-3} \cmidrule(lr){4-5} \cmidrule(lr){6-7} \cmidrule(lr){8-9}
& \textbf{Acc} & \textbf{H} & \textbf{Acc} & \textbf{H} & \textbf{Acc} & \textbf{H} & \textbf{Acc} & \textbf{H} \\
\midrule
OpenGAN & 96.25 & 92.18 & 92.34 & 91.05 & 81.15 & 81.23 & 93.42 & 90.58 \\
Cumix & 96.89 & 94.26 & 96.18 & 92.85 & 82.34 & 86.74 & 92.85 & 91.84 \\
ODG-CLIP & 99.64 & 99.78 & 98.85 & 96.74 & 86.23 & 91.15 & 96.42 & 95.31 \\
\midrule
\textbf{SGDG} & \textbf{99.90} & \textbf{99.97} & \textbf{99.50} & \textbf{98.97} & \textbf{88.70} & \textbf{92.80} & \textbf{98.87} & \textbf{97.06} \\
\bottomrule
\end{tabular}}
\end{table}

\subsection{Semantic Attention Visualization}

\begin{figure}[h]
\centering
\includegraphics[width=1\columnwidth]{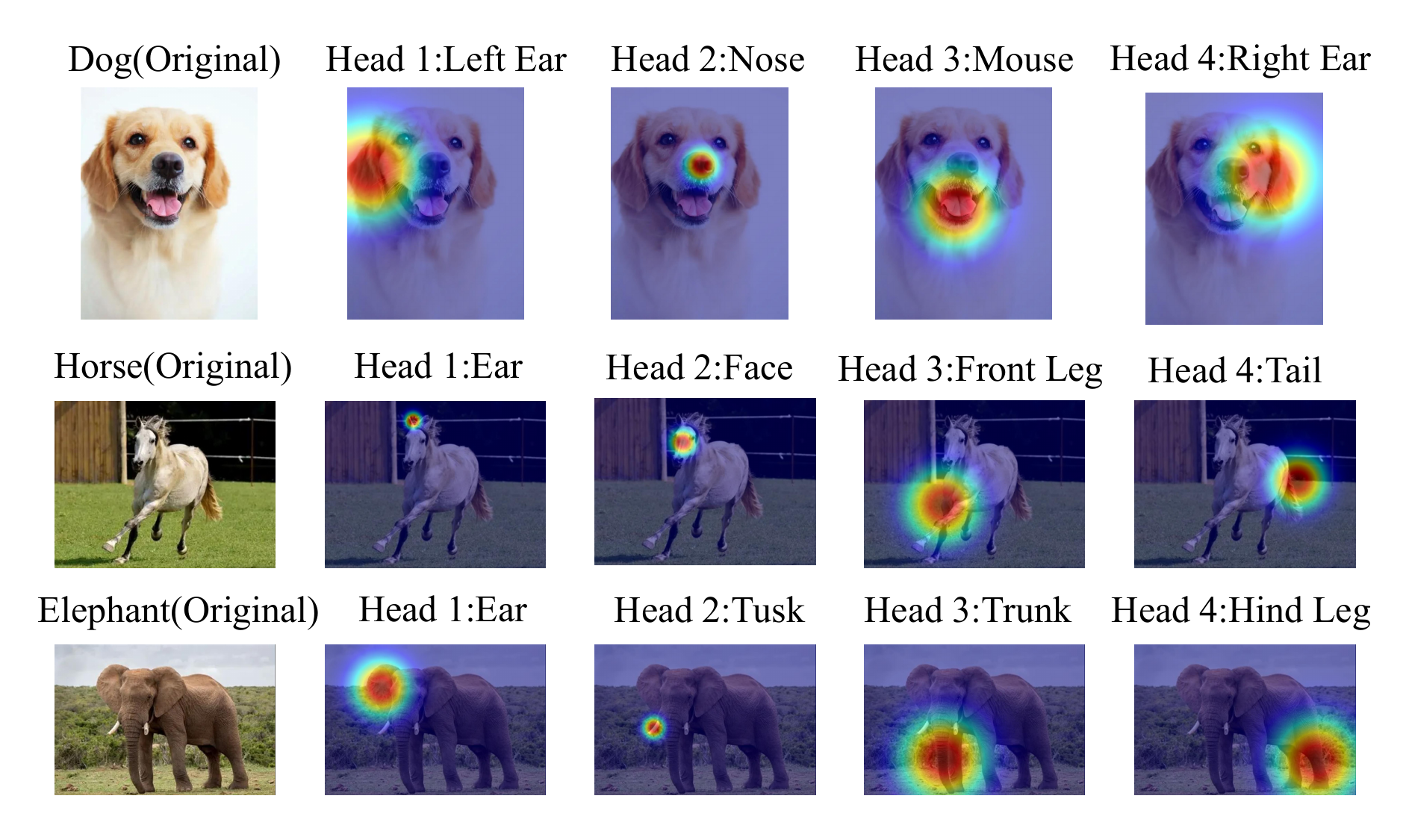}
\caption{Visualization of multi-head semantic attention across different categories. Each row represents a distinct category (Dog, Horse, Elephant), with the leftmost column showing the original image. The four attention heads (columns 2-5) focus on discriminative semantic regions.}
\label{fig:3}
\end{figure}

We visualize the learned attention distributions in Figure ~\ref{fig:3}, each of the $K$=4 attention heads automatically focuses on distinct discriminative regions without part-level supervision. For the dog, attention heads concentrate on 
ears, nose, and mouth. For the horse, heads capture ears, face, front leg, and tail. For the elephant, attention focuses on ear, tusk, trunk, and hind leg. These visualizations demonstrate that SAPE successfully extracts 
category-specific fine-grained semantic features, enabling the model to distinguish semantically similar categories and handle hard unknowns in OSDG scenarios.

\section{Conclusions}

This paper presents SeeCLIP, a semantic-enhanced framework for fine-grained open-set domain generalization. Integrating semantic-aware prompt enhancement, semantic-guided diffusion generation, and duplex contrastive learning, it effectively distinguishes semantically similar unknowns. Extensive experiments validate consistent gains over SOTA methods, establishing a semantic-prioritized paradigm for open-set recognition under distribution drift.

\section*{Acknowledgments}
This work was supported by the National Natural Science Foundation of China No. 62576174.

\bibliography{CameraReady/LaTeX/refs}

@article{wang2022generalizing,
  title={Generalizing to unseen domains: A survey on domain generalization},
  author={Wang, Jindong and Lan, Cuiling and Liu, Chang and Ouyang, Yidong and Qin, Tao and Lu, Wang and Chen, Yiqiang and Zeng, Wenjun and Yu, Philip S},
  journal={IEEE transactions on knowledge and data engineering},
  volume={35},
  number={8},
  pages={8052--8072},
  year={2022},
  publisher={IEEE}
}

@inproceedings{li2017deeper,
  title={Deeper, broader and artier domain generalization},
  author={Li, Da and Yang, Yongxin and Song, Yi-Zhe and Hospedales, Timothy M},
  booktitle={Proceedings of the IEEE international conference on computer vision},
  pages={5542--5550},
  year={2017}
}

@inproceedings{zhou2020learning,
  title={Learning to generate novel domains for domain generalization},
  author={Zhou, Kaiyang and Yang, Yongxin and Hospedales, Timothy and Xiang, Tao},
  booktitle={European conference on computer vision},
  pages={561--578},
  year={2020},
  organization={Springer}
}

@inproceedings{wang2023generalizable,
  title={Generalizable decision boundaries: Dualistic meta-learning for open set domain generalization},
  author={Wang, Xiran and Zhang, Jian and Qi, Lei and Shi, Yinghuan},
  booktitle={Proceedings of the IEEE/CVF International Conference on Computer Vision},
  pages={11564--11573},
  year={2023}
}

@article{yang2024open,
  title={Open set recognition in real world},
  author={Yang, Zhen and Yue, Jun and Ghamisi, Pedram and Zhang, Shiliang and Ma, Jiayi and Fang, Leyuan},
  journal={International Journal of Computer Vision},
  volume={132},
  number={8},
  pages={3208--3231},
  year={2024},
  publisher={Springer}
}

@article{sun2023survey,
  title={A survey on open-set image recognition},
  author={Sun, Jiayin and Dong, Qiulei},
  journal={arXiv preprint arXiv:2312.15571},
  year={2023}
}

@inproceedings{vaze2022gcd,
  title={Generalized Category Discovery},
  author={Vaze, Samarth and Han, Kai and Vedaldi, Andrea and Zisserman, Andrew},
  booktitle={Proceedings of the IEEE/CVF conference on computer vision and pattern recognition},
  pages={7492--7501},
  year={2022}
}

@inproceedings{shu2021open,
  title={Open domain generalization with domain-augmented meta-learning},
  author={Shu, Yang and Cao, Zhangjie and Wang, Chenyu and Wang, Jianmin and Long, Mingsheng},
  booktitle={Proceedings of the IEEE/CVF conference on computer vision and pattern recognition},
  pages={9624--9633},
  year={2021}
}

@inproceedings{singha2024unknown,
  title={Unknown Prompt the only Lacuna: Unveiling CLIP's Potential for Open Domain Generalization},
  author={Singha, Mainak and Jha, Ankit and Bose, Shirsha and Nair, Ashwin and Abdar, Moloud and Banerjee, Biplab},
  booktitle={Proceedings of the IEEE/CVF Conference on Computer Vision and Pattern Recognition},
  pages={13309--13319},
  year={2024}
}

@inproceedings{gupta2025osloprompt,
  title={OSLoPrompt: Bridging Low-Supervision Challenges and Open-Set Domain Generalization in CLIP},
  author={Gupta, Divyam and Singha, Mainak and Rongali, Sai Bhargav and Jha, Ankit and Khan, Muhammad Haris and Banerjee, Biplab and others},
  booktitle={Proceedings of the Computer Vision and Pattern Recognition Conference},
  pages={10110--10120},
  year={2025}
}

@inproceedings{radford2021learning,
  title={Learning transferable visual models from natural language supervision},
  author={Radford, Alec and Kim, Jong Wook and Hallacy, Chris and Ramesh, Aditya and Goh, Gabriel and Agarwal, Sandhini and Sastry, Girish and Askell, Amanda and Mishkin, Pamela and Clark, Jack and others},
  booktitle={International conference on machine learning},
  pages={8748--8763},
  year={2021},
  organization={PmLR}
}

@inproceedings{jia2021scaling,
  title={Scaling up visual and vision-language representation learning with noisy text supervision},
  author={Jia, Chao and Yang, Yinfei and Xia, Ye and Chen, Yi-Ting and Parekh, Zarana and Pham, Hieu and Le, Quoc and Sung, Yun-Hsuan and Li, Zhen and Duerig, Tom},
  booktitle={International conference on machine learning},
  pages={4904--4916},
  year={2021},
  organization={PmLR}
}

@inproceedings{cherti2023reproducible,
  title={Reproducible scaling laws for contrastive language-image learning},
  author={Cherti, Mehdi and Beaumont, Romain and Wightman, Ross and Kalantidis, Yannis and Adeli, Ehsan and Ranzato, Marc'Aurelio and Caron, Mathilde and Alayrac, Jean-Baptiste},
  booktitle={Proceedings of the IEEE/CVF conference on computer vision and pattern recognition},
  pages={2818--2829},
  year={2023}
}

@inproceedings{lang2024coarse,
  title={From coarse to fine-grained open-set recognition},
  author={Lang, Nico and Sn{\ae}bjarnarson, V{\'e}steinn and Cole, Elijah and Mac Aodha, Oisin and Igel, Christian and Belongie, Serge},
  booktitle={Proceedings of the IEEE/CVF conference on computer vision and pattern recognition},
  pages={17804--17814},
  year={2024}
}

@inproceedings{bele2024learning,
  title={Learning class and domain augmentations for single-source open-domain generalization},
  author={Bele, Prathmesh and Bundele, Valay and Bhattacharya, Avigyan and Jha, Ankit and Roig, Gemma and Banerjee, Biplab},
  booktitle={Proceedings of the IEEE/CVF Winter Conference on Applications of Computer Vision},
  pages={1816--1826},
  year={2024}
}

@inproceedings{zhou2022conditional,
  title={Conditional prompt learning for vision-language models},
  author={Zhou, Kaiyang and Yang, Jingkang and Loy, Chen Change and Liu, Ziwei},
  booktitle={Proceedings of the IEEE/CVF conference on computer vision and pattern recognition},
  pages={16816--16825},
  year={2022}
}

@article{zhou2022learning,
  title={Learning to prompt for vision-language models},
  author={Zhou, Kaiyang and Yang, Jingkang and Loy, Chen Change and Liu, Ziwei},
  journal={International Journal of Computer Vision},
  volume={130},
  number={9},
  pages={2337--2348},
  year={2022},
  publisher={Springer}
}

@article{yu2024fine,
  title={Fine-Grained Domain Generalization with Feature Structuralization},
  author={Yu, Wenlong and Chen, Dongyue and Wang, Qilong and Hu, Qinghua},
  journal={arXiv preprint arXiv:2406.09166},
  year={2024}
}

@inproceedings{bi2025learning,
  title={Learning fine-grained domain generalization via hyperbolic state space hallucination},
  author={Bi, Qi and Yi, Jingjun and Zhan, Haolan and Ji, Wei and Xia, Gui-Song},
  booktitle={Proceedings of the AAAI Conference on Artificial Intelligence},
  volume={39},
  number={2},
  pages={1853--1861},
  year={2025}
}

@inproceedings{bendale2016towards,
  title={Towards open set deep networks},
  author={Bendale, Abhijit and Boult, Terrance E},
  booktitle={Proceedings of the IEEE conference on computer vision and pattern recognition},
  pages={1563--1572},
  year={2016}
}

@inproceedings{kong2021opengan,
  title={Opengan: Open-set recognition via open data generation},
  author={Kong, Shu and Ramanan, Deva},
  booktitle={Proceedings of the IEEE/CVF international conference on computer vision},
  pages={813--822},
  year={2021}
}

@inproceedings{panareda2017open,
  title={Open set domain adaptation},
  author={Panareda Busto, Pau and Gall, Juergen},
  booktitle={Proceedings of the IEEE international conference on computer vision},
  pages={754--763},
  year={2017}
}

@article{bose2023beyond,
  title={Beyond boundaries: A novel data-augmentation discourse for open domain generalization},
  author={Bose, Shirsha and Jha, Ankit and Kandala, Hitesh and Banerjee, Biplab},
  journal={Transactions on Machine Learning Research},
  year={2023}
}

@inproceedings{rakshit2022osda,
  title={Open-set domain adaptation under few source-domain labeled samples},
  author={Rakshit, Sourya and Bandyopadhyay, Himadri and Bharambe, Prathamesh and Patel, Vishal M},
  booktitle={Proceedings of the IEEE/CVF conference on computer vision and pattern recognition},
  pages={4029--4038},
  year={2022}
}

@inproceedings{rombach2022high,
  title={High-resolution image synthesis with latent diffusion models},
  author={Rombach, Robin and Blattmann, Andreas and Lorenz, Dominik and Esser, Patrick and Ommer, Bj{\"o}rn},
  booktitle={Proceedings of the IEEE/CVF conference on computer vision and pattern recognition},
  pages={10684--10695},
  year={2022}
}

@inproceedings{wang2022dualprompt,
  title={Dualprompt: Complementary prompting for rehearsal-free continual learning},
  author={Wang, Zifeng and Zhang, Zizhao and Ebrahimi, Sayna and Sun, Ruoxi and Zhang, Han and Lee, Chen-Yu and Ren, Xiaoqi and Su, Guolong and Perot, Vincent and Dy, Jennifer and others},
  booktitle={European conference on computer vision},
  pages={631--648},
  year={2022},
  organization={Springer}
}

@inproceedings{khattak2023maple,
  title={Maple: Multi-modal prompt learning},
  author={Khattak, Muhammad Uzair and Rasheed, Hanoona and Maaz, Muhammad and Khan, Salman and Khan, Fahad Shahbaz},
  booktitle={Proceedings of the IEEE/CVF conference on computer vision and pattern recognition},
  pages={19113--19122},
  year={2023}
}

@inproceedings{zhang2025less,
  title={Less Attention is More: Prompt Transformer for Generalized Category Discovery},
  author={Zhang, Wei and Zhang, Baopeng and Teng, Zhu and Luo, Wenxin and Zou, Junnan and Fan, Jianping},
  booktitle={Proceedings of the Computer Vision and Pattern Recognition Conference},
  pages={30322--30331},
  year={2025}
}

@article{ho2022classifier,
  title={Classifier-free diffusion guidance},
  author={Ho, Jonathan and Salimans, Tim},
  journal={arXiv preprint arXiv:2207.12598},
  year={2022}
}

\end{document}